%% file: main.tex
\documentclass[11pt,a4paper]{article}
\usepackage[hyperref]{emnlp-ijcnlp-2019}
\usepackage{times,amsmath}
\usepackage{latexsym}
\usepackage{graphicx}
\usepackage{url}
\usepackage{todonotes}
\usepackage{multirow}
\usepackage{enumitem}
\DeclareMathOperator*{\argmax}{arg\,max}

\aclfinalcopy 

\title{Multi-label Categorization of Accounts of Sexism using a Neural Framework}

\author{Pulkit Parikh$^1$, Harika Abburi$^1$, Pinkesh Badjatiya$^1$, Radhika Krishnan$^1$, \\\textbf{Niyati Chhaya}$^{1,2}$,
\textbf{Manish Gupta}$^{1*}$, \textbf{Vasudeva Varma}$^1$ \\
$^1$IIIT-Hyderabad, India \\
$^2$Adobe Research, India\\
{\{pulkit.parikh,harika.a,pinkesh.badjatiya\}@research.iiit.ac.in},\\ radhika.krishnan@iiit.ac.in, nchhaya@adobe.com, \{manish.gupta,vv\}@iiit.ac.in}

\date{}

\makeatletter
\g@addto@macro\normalsize{%
  \setlength\abovedisplayskip{-10pt}
  \setlength\belowdisplayskip{2pt}
  \setlength\abovedisplayshortskip{-10pt}
  \setlength\belowdisplayshortskip{2pt}
}
\makeatletter
\def\blfootnote{\gdef\@thefnmark{}\@footnotetext}
\makeatother
\begin{document}
\maketitle
\begin{abstract}
Sexism, an injustice that subjects women and girls to enormous suffering, manifests in blatant as well as subtle ways. In the wake of growing documentation of experiences of sexism on the web, the automatic categorization of accounts of sexism has the potential to assist social scientists and policy makers in studying and countering sexism better. The existing work on sexism classification, which is different from sexism detection, has certain limitations in terms of the categories of sexism used and/or whether they can co-occur. To the best of our knowledge, this is the first work on the multi-label classification of sexism of any kind(s), and we contribute the largest dataset for sexism categorization. We develop a neural solution for this multi-label classification that can combine sentence representations obtained using models such as BERT with distributional and linguistic word embeddings using a flexible, hierarchical architecture involving recurrent components and optional convolutional ones. Further, we leverage unlabeled accounts of sexism to infuse domain-specific elements into our framework. The best proposed method outperforms several deep learning as well as traditional machine learning baselines by an appreciable margin.
\end{abstract}
\blfootnote{$^*$The author is also an applied researcher at Microsoft.}
\input{intro}
\input{related_work}
\input{dataset.tex}
\input{approach}
\input{experiments}

\input{conclusion}
\newpage
\bibliography{emnlp-ijcnlp-2019}
\bibliographystyle{acl_natbib}

\end{document}

%% file: intro.tex
\section{Introduction}
Sexism, discrimination on the basis of one's sex, prevails in our society in numerous forms, causing immense suffering to women and girls. Online forums have enabled victims of sexism to share their experiences freely and widely by facilitating anonymity and connecting far-away people. A meaningful categorization of these accounts of sexism can play a part in analyzing sexism with a view to developing sensitization programs, systemic safeguards, and other mechanisms against this injustice. Given the substantial volume of posts related to sexism on digital media, automated sexism categorization can aid social scientists and policy makers in combating sexism by conducting such analyses efficiently.

While sexism is detected as a category of hate in some of the hate speech classification work~\cite{badjatiya2017hate, waseem2016hate}, it does not perform sexism classification. Except the work on categorizing sexual harassment by~\citet{karlekar2018harass}, the prior work on classifying sexism assumes the categories to be mutually exclusive~\cite{anzovino2018sexism, jha2017sexism}. Moreover, the existing category sets number between 2 to 5. In this paper, we focus on the new problem of the multi-label categorization of an account of sexism reporting any type(s) of sexism. We create a dataset comprising 13023 accounts of sexism, including first-person accounts from survivors, each tagged with at least one of $23$ categories of sexism. The categories were defined keeping in mind the discourse and campaigns on gender-related issues along with potential policy implications, under the guidance of a social scientist. Ten annotators, most of whom have formally studied topics related to gender and/or sexuality, were recruited to label textual accounts of sexism. The accounts are drawn from the Everyday Sexism Project website\footnote{\url{https://everydaysexism.com}}, where voluntary contributors from all over the world document experiences of sexism suffered or witnessed by them. For classification experiments, the categories found in less than $400$ accounts in our dataset are appropriately merged with others, resulting in $14$ categories.



The rationale for formulating this classification as multi-label is that many experiences inherently involve multiple types of sexism. For instance, \textit{``I overheard a co-worker saying that I should be in more team events and photos because I am pleasing to the eye! Disgusting."} describes an experience of sexism wherein the victim was subjected to three types of sexism, namely hyper-sexualization, sexual harassment, and hostile work environment.




We develop a novel neural architecture for the multi-label classification of accounts of sexism that enables flexibly combining sentence representations created using models such as BERT (Bidirectional Encoder Representations from Transformers)~\cite{bert} with distributional word embeddings like ELMo (Embeddings from Language Models)~\cite{elmo} and Global Vectors (GloVe)~\cite{glove} and a linguistic feature representation through hierarchical convolutional and/or recurrent operations. Leveraging general-purpose models such as BERT for encoding sentences likely makes our model better equipped to capture semantic aspects effectively, since they are trained on substantially larger textual data than the domain-specific labeled data that we have. Moreover, we adapt a BERT model for the domain of instances of sexism using unlabeled data. Embeddings from sentence encoders are complemented by sentence representations built from word embeddings as a function of trainable neural network parameters. We explore multiple ways to deal with the multi-label aspect. The adopted method produces label-wise probabilities directly and simultaneously using shared weights and a joint loss function. Our experimentation finds multiple instances of the proposed framework outperforming several diverse baselines on established multi-label classification metrics.


Our key contributions are summarized below.
\begin{itemize}[leftmargin=*]
\item We propose a neural framework for the multi-label classification of accounts of sexism that can combine sentence representations built from word embeddings of different kinds through learnable model parameters with those created using pre-trained models. It yields results superior to many deep learning and traditional machine learning baselines.
\item To the best of our knowledge, this is the first work on classifying an account recounting any type(s) of sexism without the assumption of the mutual exclusivity of classes.
\item We provide a dataset consisting of $13023$ accounts of sexism by survivors and observers annotated with one or more of $23$ carefully formulated categories of sexism.
\end{itemize}


%% file: related_work.tex
\section{Related Work}
\label{sec:relatedwork}
Substantial work has been directed to hate speech detection in recent years. Since some of it involves the detection of sexism \cite{badjatiya2017hate,waseem2016hate}, we review it along with the work on sexism classification.
\subsection{Hate Speech Detection}
\citet{warner2012hate} identify anti-semitic hate speech using SVM. \citet{gao2017hate} perform hate speech detection in a weakly supervised fashion. \citet{nobata2016hate} distinguish abusive comments from clean ones through various NLP and embedding-derived features. \citet{burnap2016hate} classify cyber hate with respect to race, disability, and sexual orientation using text parsing to extract typed dependencies. \citet{waseem2016hate} explore the role of extra-linguistic features along with character n-grams in classifying tweets as racist, sexist or neither. \citet{badjatiya2017hate} experiment with various deep learning approaches for the same three-way classification. \citet{zhang2018hate} explore skipped CNN and a combination of CNN and GRU for hate speech detection. Unlike these papers, we seek to categorize accounts of sexism, a specific form of discrimination or hate.

\subsection{Sexism Categorization}
 \citet{anzovino2018sexism} use features such as (Part of Speech) n-grams and text embeddings for the 5-class categorization of misogynistic language. \citet{jafarpour2018sexism} classify sexist tweets into one of four categories, which deal with harassment and threats. \citet{jha2017sexism} experiment with SVM, biLSTM with attention, and fastText to categorize tweets as benevolent, hostile, or non-sexist. Their way of categorizing sexism relates to the manner in which sexism is stated. While it is applicable to all sexist remarks, narrated accounts of sexism may not always capture how the perpetrators stated sexism. \citet{karlekar2018harass} focus on accounts of sexual harassment, exploring CNN and/or RNN for their 3-class classification. As far as we know, our paper presents the first attempt to categorize accounts involving any type(s) of sexism in a multi-label way. Moreover, we provide a larger dataset and significantly more extensive or finer-grained categorization scheme than these papers.

%% file: dataset.tex
\section{Dataset Construction}
\label{sec:dataset}
\setlength{\tabcolsep}{4pt}
\begin{table*}[!t]
\centering
\scriptsize
\caption{Descriptions of the categories of sexism used in our dataset}\label{dataset}
\vspace{-10pt}
\begin{tabular}{|p{1.25in}|p{4.55in}|} 
\hline
Category & Description\\
\hline
\hline
Role stereotyping & Socially constructed false generalizations about certain roles being more appropriate for women; also applies to such misconceptions about men\\
\hline
Attribute stereotyping & Mistaken linkage of women with some physical, psychological, or behavioral qualities or likes/dislikes; also applies to such false notions about men\\
\hline
Body shaming & Objectionable comments or behaviour concerning appearance including the promotion of certain body types or standards\\
\hline
Hyper-sexualization (excluding body shaming) & Unwarranted focus on physical aspects or sexual acts\\
\hline
Internalized sexism & The perpetration of sexism by women via comments or other actions\\
\hline
Pay gap & Unequal salaries for men and women for the same work profile\\
\hline
Hostile work environment (excluding pay gap) & Sexism encountered by an employee at the workplace; also applies when a sexist misdeed committed outside the workplace by a co-worker makes working uncomfortable for the victim\\
\hline
Denial or trivialization of sexist misconduct & Denial or downplaying of sexist wrongdoings\\
\hline
Threats & All threats including wishing for violence or joking about it, stalking, threatening gestures, or rape threats\\
\hline
Rape & FBI's expanded definition of rape\\
\hline
Sexual assault (excluding rape) & Any sexual contact without consent; unwanted touching\\
\hline
Sexual harassment (excluding assault) & Any sexually objectionable behaviour\\
\hline
Tone policing & Comments or actions that cause or aggravate restrictions on how women communicate\\
\hline
Moral policing (excluding tone policing) & The promotion of discriminatory codes of conduct for women in the guise of morality; also applies to statements that feed into such codes and narratives\\
\hline
Victim blaming & The act of holding the victim responsible (fully or partially) for sexual harassment, violence, or other sexism perpetrated against her\\
\hline
Slut shaming & Inappropriate comments made about women 1) deviating from conservative expectations relating to sex or 2) dressing in a certain way when it gets linked to sexual availability\\
\hline
Motherhood-related discrimination & Shaming, prejudices, or other discrimination or misconduct related to the notion of motherhood; also applies to the violation of reproductive rights\\
\hline
Menstruation-related discrimination & Shaming, prejudices, or other discrimination or wrongdoings related to periods\\
\hline
Religion-based sexism & Sexist discrimination or prejudices stemming from religious scriptures or constructs\\
\hline
Physical violence (excluding sexual violence) & Domestic abuse, murder, kidnapping, confinement, or other physical acts of violence linked to sexism\\
\hline
Mansplaining & A woman being condescendingly talked down to by a man; also applies when a man gives an unsolicited advice or explanation to a woman related to something she knows well that she disapproves of\\
\hline
Gaslighting & Sexist manipulation of the victim through psychological means into doubting her own sanity\\
\hline
Other & Any type of sexism not covered by the above categories\\
\hline
\end{tabular}
\end{table*}
 
Creating our multi-label sexism account categorization dataset entailed two parts: textual data collection and data annotation. To collect data, we crawled the Everyday Sexism Project website, which receives numerous accounts of sexism from survivors themselves as well as observers. After removing entries with less than $7$ words, around 20000 entries were shortlisted for annotation; we prioritized shorter ones and tried to approximate the tag distribution on the website. Though shorter entries were preferred keeping in mind the potential future work of transfer learning to Twitter content, our neural framework is devised in a size-agnostic way.

Under the direction of a social scientist, $23$ categories of sexism were formulated taking into account gender-related discourse and campaigns~\cite{dutta2013india, eccles1990gender, mead1963sex, menon2012seeing} as well as possible impact on public policy. Table~\ref{dataset} provides succinct descriptions for the categories.
  
\begin{figure}[!t]
\centering
\includegraphics[width=0.7\columnwidth]{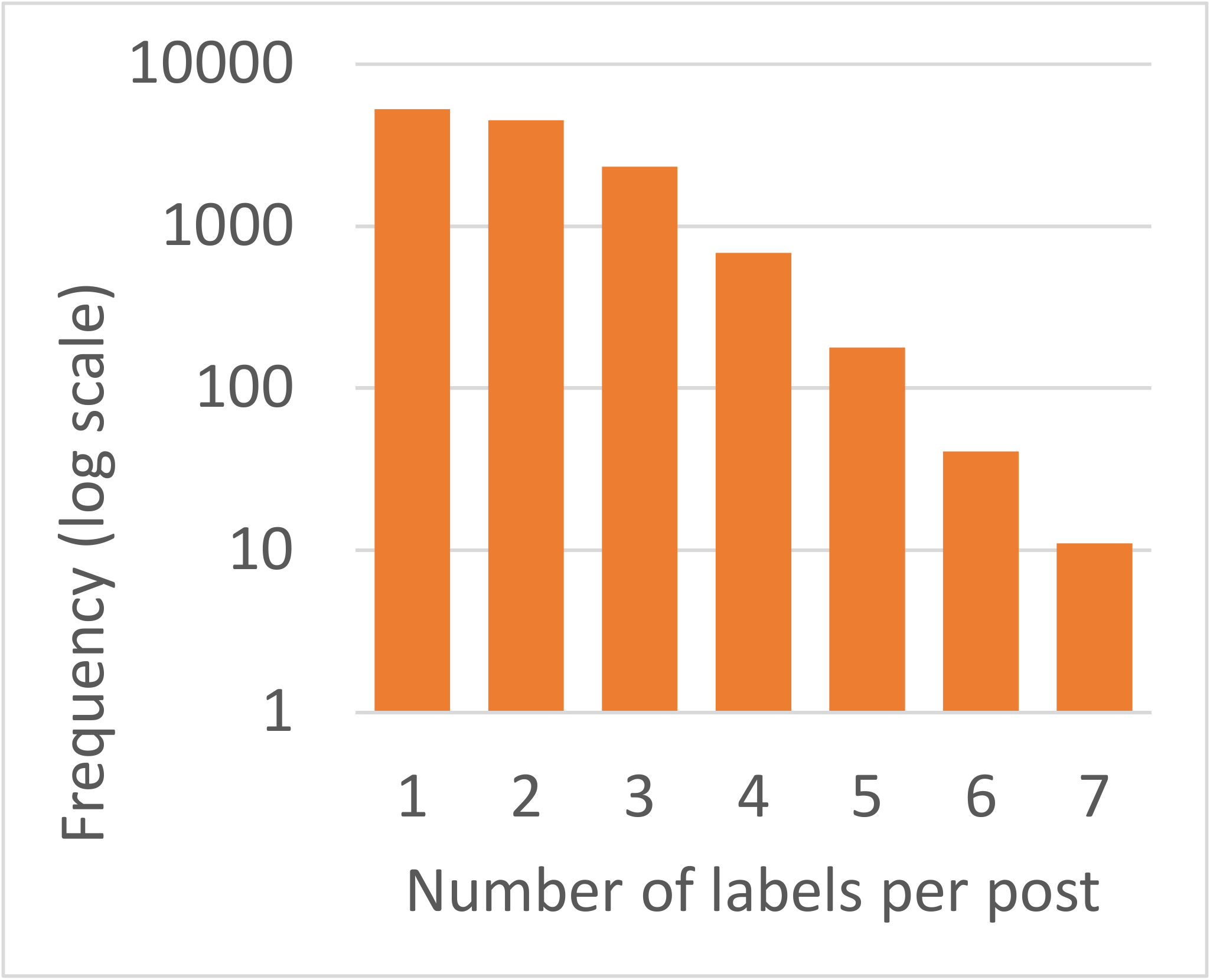}%
\caption{Frequency distribution of \#labels per post}\label{fig:histogram}
\end{figure}

We followed a three-phase annotation process to ensure that the categorization of each account of sexism in the final dataset involved the labeling of it by at least two of our $10$ annotators, most of whom had studied topics related to gender and/or sexuality formally. The annotators were given detailed guidelines, which evolved during the course of their work. Each annotator was given training, which included a pilot round involving evaluation and feedback. Phase $1$ involved identifying one or more textual portions containing distinct accounts of sexism from an entry obtained from Everyday Sexism Project and subsequently tagging each portion with at least one of the $23$ categories of sexism, producing over $23000$ labeled accounts. In phases $2$ and $3$, we sought redundancy of annotations for improved quality, as permitted by the availability of annotators adequately knowledgeable about sexism. Over $21000$ accounts were categorized again in phase $2$ such that the annotators for phases $1$ and $2$ were different. The inter-annotator agreement across phases $1$ and $2$, measured by the average of the Cohen's Kappa~\cite{kappa1, kappa2} scores for the per-category pairs of binary label vectors, is $0.584$. Each account for which the label sets annotated across phases 1 and 2 were identical was included in the dataset along with the associated label set. In phase $3$, some of the accounts for which there was a mismatch between the phase $1$ and phase $2$ annotations were selected. For each account, the annotators were presented with only the mismatched categories and asked to select or reject each. Duplicates and records for which the Everyday Sexism Project entry numbers match but the accounts do not fully match were removed at multiple stages. In order to improve the annotation reliability further, some records for which the annotations differed across phases $1$ and $2$ were discarded based on the annotators involved and sensitivities of the categories, resulting in a multi-label sexism categorization dataset of $13023$ accounts. For our automated sexism classification experiments, we merge the categories found in less than $400$ records with others as follows, resulting in $14$ categories. `Menstruation-related discrimination' and `Motherhood-related discrimination' are merged into `Motherhood and menstruation related discrimination'; `Mansplaining', `Gaslighting', `Religion-based sexism', `Physical violence (excluding sexual violence)', and `Other' are merged into `Other'; `Pay gap' and `Hostile work environment (excluding pay gap)' are merged into `Hostile work environment'; `Tone policing', `Moral policing (excluding tone policing)', and `Victim blaming' are merged into `Moral policing and victim blaming'; `Rape' and `Sexual assault (excluding rape)' are merged into `Sexual assault'. Our dataset, however, retains all $23$ categories. Fig.~\ref{fig:histogram} shows the frequency distribution of the number of labels per account in the dataset, demonstrating the multi-label nature of instances of sexism.



\textbf{Caution}: 1) The category frequencies in our dataset (used for merging categories) do not represent the real-world frequencies of those categories of sexism, as they are affected by several factors including the bias of our sampling scheme toward smaller posts and the small size of our dataset compared to the immense degree of prevalence of sexism in the world. 2) Labeling categories of sexism can be complex in many cases. Hence, despite our best efforts, our labeled data may contain inaccuracies or discrepancies. We also recognize that our categorization scheme could be improved.

\subsection{Ethical Data Use and Release}

We are committed to following ethical practices, which includes protecting the privacy and anonymity of the victims. We only use accounts of sexism and tags from entries on the Everyday Sexism Project website (ESP). The entry titles, which could contain sensitive information related to the names or locations of the victims (or contributors), are not saved or used at all.

Our dataset can be requested for academic purposes only by providing some prerequisites as recommended by an ethics committee and agreeing to certain terms through our website\footnote{\url{https://irel.iiit.ac.in/sexism-classification}}. The requesters who fulfill these conditions will be emailed 1) the data comprising only numerical placeholders and labels, 2) a script that fetches only accounts of sexism from ESP to obtain the account for each placeholder, and 3) the annotation guidelines used. We have devised this method to ensure that if an entry gets removed from ESP by a victim (or contributor), any and all parts of it in our dataset will also be removed.

%% file: approach.tex
\section{Sexism Categorization Approach}
\label{sec:approach}
\begin{figure}[!t]
\includegraphics[width=\columnwidth]{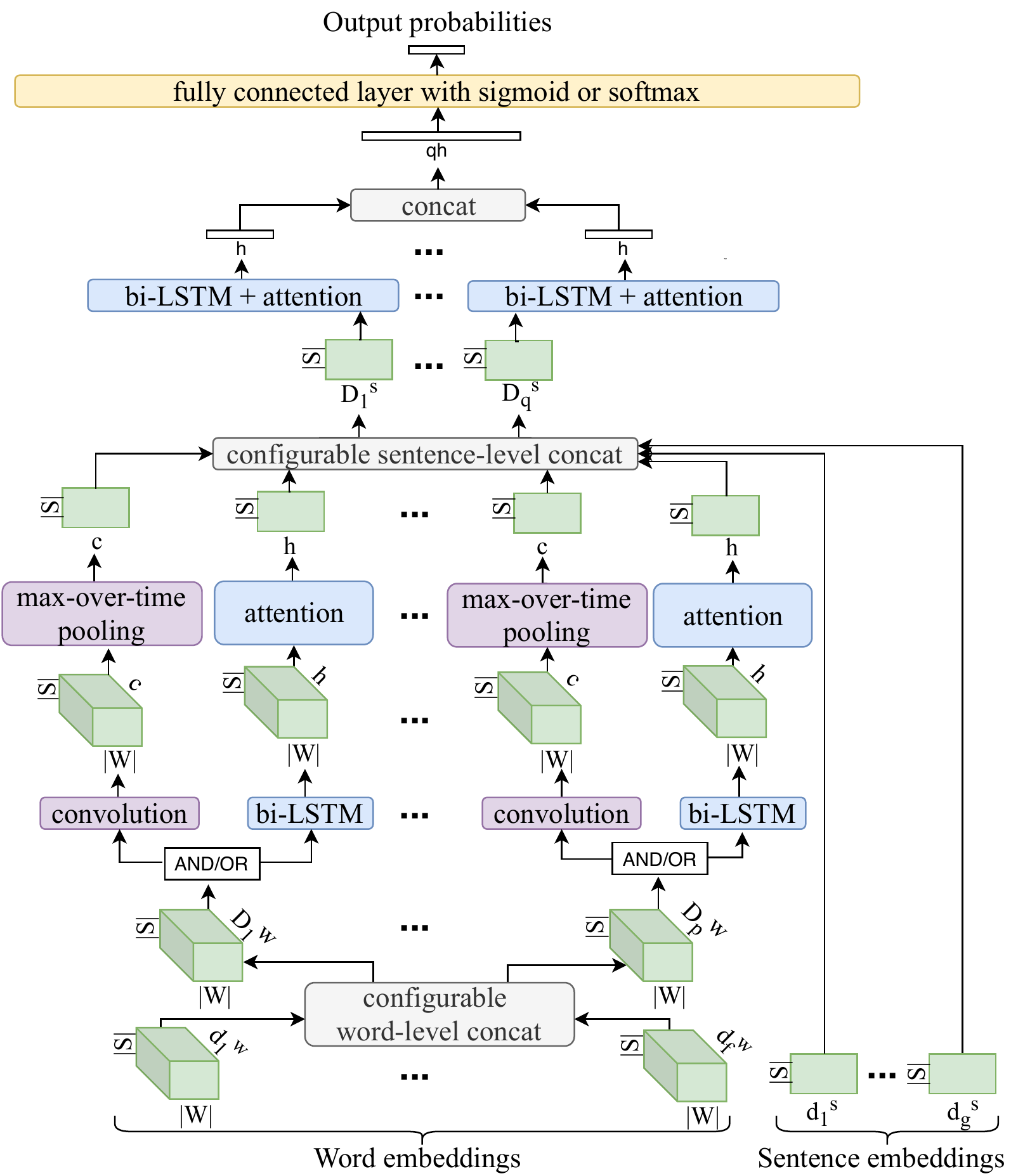}%
\caption{Proposed sexism categorization architecture}\label{fig_archi}
\end{figure}

Given an account of sexism (post), our objective is to predict a list of up to $14$ applicable categories of sexism, making this a multi-label multi-class classification task. In this section, we detail our proposed framework, which enables combining sentence representations derived from word embeddings using trainable model parameters with those obtained using general-purpose models. Our architecture is depicted in Fig.~\ref{fig_archi}. We also discuss how we tap unlabeled data and loss functions.

\subsection{Sexism Categorization Architecture}
Let each post contain a maximum of $|S|$ sentences with a maximum of $|W|$ words per sentence. Every word (or sentence) can be represented using multiple word (or sentence) embedding methods. Let $f$ (or $g$) be the number of word (or sentence) embedding methods chosen. Let $d_i^w$ (or $d_j^s$) be the embedding dimension for the $i^{th}$ (or $j^{th}$) word (or sentence) embedding scheme. Each post is represented using two kinds of tensors: (a) $f$ tensors $\in \mathcal{R}^{|S| \times |W| \times d_i^w}$ created using different word embeddings, and (b) $g$ tensors $\in \mathcal{R}^{|S| \times d_j^s}$ constructed using different sentence encoders.

First, subsets of the $f$ tensors based on word embeddings are concatenated in a configurable manner (\textit{configurable word-level concat} in Fig.~\ref{fig_archi}), producing $p$ tensors $\in \mathcal{R}^{|S| \times |W| \times D_i^w}$, where $D_i^w$ is the dimension resulting from the $i^{th}$ concatenation. Next, we construct vector representations for the sentences word-embedded in each of the $p$ tensors using CNN-based and/or LSTM-based operations as configured. The CNN-based operations begin with convolutional filters being applied along the word dimension~\cite{kim2014cnn} to generate many bigram, trigram and 4-gram based features. This is followed by max-over-time pooling, which picks the largest value for each filter and produces a sentence representing tensor $\in \mathcal{R}^{|S| \times c}$, where $c$ is the total number of convolutional filters used. The LSTM-based components include biLSTM followed by an attention mechanism~\cite{yang2016attention} through which the LSTM outputs across time steps are aggregated into a vector representation for each sentence, resulting in a tensor $\in \mathcal{R}^{|S| \times h}$, where $h$ is the bi-LSTM output length. At this stage, we have three types of sentence representing tensors if both CNN-based and RNN-based operations are chosen to be applied on all word embedding tensors: (a) $p$ tensors $\in \mathcal{R}^{|S| \times c}$ from the CNN-based processing, (b) $p$ tensors $\in \mathcal{R}^{|S| \times h}$ from the LSTM-based processing, and (c) $g$ tensors $\in \mathcal{R}^{|S| \times d_j^s}$ obtained using general-purpose sentence encoders. 

From these sentence representing tensors, subsets are concatenated to produce $q$ tensors $\in \mathcal{R}^{|S| \times D_j^s}$ (\textit{configurable sentence-level concat} in Fig.~\ref{fig_archi}), where $D_j^s$ is the dimension stemming from the $j^{th}$ concatenation. The sequence of sentence vectors in each of these $q$ tensors is then passed through bi-LSTM followed by attention-based aggregation, producing $q$ representations for a post collectively. These vectors are then concatenated to produce the overall post representation. The final step involves a fully connected layer with a sigmoid or softmax non-linearity depending on the loss function used, generating the output probabilities.

\subsection{Word and Sentence Representations}
We model a post using both word embeddings and sentence embeddings. We experiment with three distributional word vectors, namely ELMo~\cite{elmo}, GloVe~\cite{glove}, and fastText~\cite{fasttext_word_embeddings}, and a linguistic feature vector. Our linguistic feature representation comprises a variety of features, namely features from the biased language detection work (assertive verb, implicative verb, hedges, factive verb, report verb, entailment, strong subjective, weak subjective, positive word, and negative word)~\cite{linguistic}, PERMA (Positive Emotion, Engagement, Relationships, Meaning, and Accomplishments) features for both polarities~\cite{perma}, associations with eight basic emotions (anger, fear, anticipation, trust, surprise, sadness, joy, and disgust) and two sentiments (negative and positive) from the NRC emotion lexicon~\cite{nrc}, and affect (valence, arousal, and dominance) scores~\cite{vad-acl2018}. Missing values are filled with zero for binary features and with the mean for non-binary ones.

We explore the following for creating sentence embeddings: BERT~\cite{bert}, Universal Sentence Encoder (USE)~\cite{use}, and InferSent~\cite{infersent}. Our choice of utilizing these models is warranted by the fact that the corpora that they are trained on are considerably bigger than the textual data that we have for supervised learning and hence likely contain greater semantic diversity. 

\subsection{Utilizing Unlabeled Data}
Models such as BERT are not trained to generate representations tuned to a specific domain. We use over $90000$ entries crawled from Everyday Sexism Project's website to tailor a pre-trained BERT model for obtaining more effective representations for our model. After removing the unlabeled entries corresponding to the posts in the test and validation data, we use the rest to tune the BERT parameters using its masked language modeling and next sentence prediction tasks. We henceforth refer to this refined model as tBERT (tuned BERT).

\subsection{Loss Function Choice}

Since the popular cross-entropy loss is inapt for our multi-label classification task in its standard form, we explore two alternatives. Binary (multi-hot) target vectors are used for both.

\subsubsection{Extended Binary Cross Entropy Loss}
We adopt an \underline{E}xtended version of the \underline{B}inary \underline{C}ross \underline{E}ntropy loss ($EBCE$), formulated as a weighted mean of label-wise binary cross entropy values in order to neutralize class imbalance.

\begin{eqnarray}
\nonumber L_{EBCE} = - \frac{1}{n} \sum_{i=1}^{n} \frac{1}{L} \sum_{j=1}^{L} w_{jy_{ij}} \left \{ y_{ij} \log(\hat p_{ij}^{\sigma}) \right.\\
\left.+ (1 - y_{ij}) \log(1 -\hat p_{ij}^{\sigma})   \right \}
\label{bin_loss}
\end{eqnarray}

Here, $n$ and $L$ denote the number of samples (posts) and the number of classes respectively. $y_{ij}$ is $1$ if label $l_j$ applies to post $x_i$ and $0$ otherwise. $\hat p_{ij}^{\sigma}$ is the estimated probability of label $l_j$ being applicable to post $x_i$ computed using a sigmoid activation atop the fully connected layer with $L$ output units. The weights for correcting class imbalance $w_{jv}$ are computed as follows.

\begin{eqnarray}
w_{jv} = \frac{n}{2|\{x_i \mid y_{ij} = v, 1 \leq i \leq n\}|}
\label{bin_w}
\end{eqnarray}

\subsubsection{Normalized Cross Entropy Loss}
We also experiment with a \underline{N}ormalized variant of the \underline{C}ross \underline{E}ntropy loss tailored for a multi-label problem configuration also mitigating class imbalance (referred to as NCE).

\begin{eqnarray}
L_{NCE} = - \frac{1}{n} \sum_{i=1}^{n} \frac{1}{|\mathbf{y_i^+}|} \sum_{j=1}^{L} w^c_j\{y_{ij}\log(\hat p_{ij})\}
\label{cross_loss}
\end{eqnarray}

Here, $\mathbf{y_i^+}$ is the set of labels applicable to post $x_i$. $\hat p_{ij}$ denotes the estimated probability of label $l_j$ being applicable to post $x_i$ computed through a softmax function. The class imbalance negating weights $w^c_j$ are generated as follows.

\begin{eqnarray}
w^c_j = \frac{n}{\sum_{i=1}^{n} \frac{y_{ij}}{|\mathbf{y_i^+}|}}
\label{cross_w}
\end{eqnarray}

Unlike in single-label multi-class classification, wherein $\argmax$ can be applied to the probability vector generated by softmax to make the prediction, we could apply a threshold on probability vector $\mathbf{\hat p_{i}}$ to predict (potentially) multiple classes for post $x_i$. Instead of using a fixed, manually tuned threshold-related parameter, we devise an automated method for estimating a per-sample cut-off position. For each sample, we descendingly sort the probability scores, compute the differences between successive (sorted) score pairs, find the index $m$ corresponding to the maximum value in the list of differences, and select the classes corresponding to indices $[1 .. m]$. Note that when sigmoid (along with $EBCE$ loss) is used instead of softmax, the prediction is made by rounding the probability vector, since it comprises the class-wise binary prediction probabilities.
\subsubsection{Discussion on Single-label Transformations}

Traditional approaches to multi-label classification include transforming the problem to one or more single-label classification problems. The Label Powerset (LP) method~\cite{boutell2004lp} treats each distinct combination of classes existing in the training set as a separate class. The standard cross-entropy loss can then be used along with softmax. This transformative method may impose a greater computational cost than the direct approach using the $EBCE$ loss since the cardinality of the transformed label set may be relatively high. Moreover, LP does not generalize to label combinations not covered in the training set. Another approach based on problem transformation is binary relevance (BR)~\cite{boutell2004lp}. An independent binary classifier is trained to predict the applicability of each label in this method. This entails training a total of $L$ classifiers, making BR computationally very expensive. Additionally, its performance is affected by the fact that it disregards correlations existing between labels. 


%% file: experiments.tex
\section{Experiments}
\label{sec:experiments}

\begin{table*}[!htb]
\setlength{\tabcolsep}{2.5pt}
\centering
\scriptsize
\caption{Results with traditional machine learning (Label Powerset)}\label{results1}
\vspace{-10pt}
\begin{tabular}{|l|c|c|c|c|c|c|c|c|c|c|c|c|c|c|c|c|} 
\hline
Features$\longrightarrow$&\multicolumn{4} {c|} {Word n-grams}&\multicolumn{4} {c|} {Character n-grams}&\multicolumn{4} {c|} {Averaged ELMo vectors}&\multicolumn{4} {c|} {Composite features}\\
\hline
Classifier$\downarrow$& $F_I$ & $F_{macro}$ & $Acc_I$ & $F_{micro}$& $F_I$ & $F_{macro}$ & $Acc_I$ & $F_{micro}$& $F_I$ & $F_{macro}$ & $Acc_I$ & $F_{micro}$& $F_I$ & $F_{macro}$ & $Acc_I$ & $F_{micro}$\\
\hline
\hline
SVM& 0.448 & 0.373 & 0.324 & 0.410& 0.449 & 0.374 & 0.331 & 0.416& 0.546 & 0.430 & 0.431 & 0.500& 0.178 & 0.094 & 0.116 & 0.174 \\
\hline
LR& 0.357 & 0.315 & 0.236 & 0.349& 0.357 &0.311  & 0.230& 0.352& \textbf{0.595} & \textbf{0.479} & \textbf{0.478}&\textbf{0.549}& 0.438 &0.370  & 0.311 & 0.421\\
\hline
RF& 0.531 & 0.398   &0.438 &0.476& 0.395 & 0.205 &0.325 & 0.349& 0.375 &0.164 &0.305 &0.331 & 0.460 & 0.311 & 0.380 & 0.415\\
\hline
\end{tabular}
\end{table*}

We evaluate the proposed framework against several baselines and provide qualitative and quantitative analyses. Our code is available on GitHub\footnote{\url{https://github.com/pulkitparikh/sexism_classification}}. Our implementation utilizes parts of the code from~\cite{agrawal2018hate, pattisapu2017medical, att_code} and libraries Keras and Scikit-learn~\cite{scikit-learn}. We reserve 15\% of the data for testing and validation each. 

\subsection{Evaluation Metrics} 
\label{subsec:eval}
Owing to the multi-label nature of this classification, standard metrics used in single-label multi-class classification are unsuitable. We adopt established example (instance) based metrics, namely F1 ($F_I$) and accuracy ($Acc_I$), and label-based metrics, namely F1 macro ($F_{macro}$) and F1 micro ($F_{micro}$) used in multi-label classification~\cite{zhang2014review}.

\subsection{Baselines} 
\label{subsec:baselines}
\noindent\underline{Random}

Labels are selected randomly as per their normalized frequencies in the training data for each test sample. 

\noindent\underline{Traditional Machine Learning (ML)}

We experiment with Support Vector Machine (SVM), Random Forests (RF), and Logistic Regression (LR). The features explored include TF-IDF on character n-grams ($1$-$5$ characters), TF-IDF on word unigrams and bigrams, the mean of the ELMo vectors for the words in a post, and the composite set of features similar to ~\citep{anzovino2018sexism} comprising n-gram based, POS-based, and doc2vec~\cite{doc2vec} features, the post length, and the adjective count.

\noindent\underline{LSTM-based Architectures}

\textbf{biLSTM}: The word embeddings for all words in a post are fed to bidirectional LSTM.

\textbf{biLSTM-Attention}: Same as biLSTM but with the attention mechanism by~\citet{yang2016attention}.

\textbf{Hierarchical-biLSTM-Attention}: For the words in each sentence, the word embeddings are passed through biLSTM with attention to create a sentence embedding. These sentence embeddings are in turn fed to another instance of biLSTM with attention. This broadly follows the architecture proposed for document classification by~\citet{yang2016attention} with GRUs replaced with LSTMs.

\textbf{Sentence embeddings with biLSTM-Attention}: Sentence representations generated using a generic encoder (BERT using bert-as-service~\cite{bertasservice}, USE, and InferSent) are passed through biLSTM with attention. 

\noindent\underline{CNN and CNN-LSTM based Architectures}

\textbf{CNN-Kim}: Similar to~\cite{kim2014cnn}, this involves applying convolutional filters followed by max-over-time pooling to the word vectors for a post.

\textbf{C-biLSTM}: In this variant of the C-LSTM architecture~\cite{clstm} somewhat related to an approach used by \citet{karlekar2018harass} for multi-label sexual harassment classification, after applying convolution on the word vectors for a post, the feature maps are stacked along the filter dimension to create a sequence of window vectors, which are then fed to biLSTM.

\textbf{CNN-biLSTM-Attention}: For each sentence, convolutional and max-over-time pooling layers are applied on the embeddings of its words. The resultant sentence representations are put through bi-LSTM with the attention mechanism. This approach is similar to~\cite{cnnlstm} with the attention scheme from~\citep{yang2016attention} added.

The architectures of the deep learning baselines have a fully connected layer with the sigmoid or softmax non-linearity (depending on the loss function used) at the end.

\subsection{Results} 
\label{subsec:results}

\begin{table}[!b]
\setlength{\tabcolsep}{1pt}
\centering
\scriptsize
\caption{Results for the proposed methods and deep learning baselines (using ELMo embeddings) with the EBCE loss and Random}\label{results2}
\vspace{-10pt}
\begin{tabular}{|c|p{1.6in}|c|c|c|c|} 
\hline
&Approach & $F_I$ & $F_{macro}$ & $Acc_I$ & $F_{micro}$\\
\hline
\hline
\multirow{10}{*}{\rotatebox{90}{Baselines}}&Random & 0.042 & 0.141 & 0.027 & 0.193\\
\cline{2-6}
&biLSTM& 0.697 & 0.616 & 0.563 & 0.658\\
\cline{2-6}
&biLSTM-Attention& 0.728 & 0.650 & 0.601 & 0.688\\
\cline{2-6}
&Hierarchical-biLSTM-Attention& 0.725 & 0.650 & 0.604 & 0.688\\
\cline{2-6}
&BERT-biLSTM-Attention& 0.656 & 0.555 & 0.502 & 0.611\\
\cline{2-6}
&USE-biLSTM-Attention& 0.628 & 0.549 & 0.468 & 0.594\\
\cline{2-6}
&InferSent-biLSTM-Attention& 0.418 & 0.37 & 0.274 & 0.399\\
\cline{2-6}
&CNN-biLSTM-Attention& 0.714 & 0.628 & 0.586 & 0.671\\
\cline{2-6}
&CNN-Kim& 0.701 & 0.622 & 0.574 & 0.669\\
\cline{2-6}
&C-biLSTM& 0.708 & 0.631 & 0.583 & 0.674\\
\hline
\multirow{8}{*}{\rotatebox{90}{Proposed methods\ \ \ \ \ \  }}&tBERT-biLSTM-Attention& 0.688 & 0.589 & 0.539 & 0.644\\
\cline{2-6}
&s(wl(ELMo), tBERT)& 0.747 & 0.675 & 0.628 & 0.710\\
\cline{2-6}
&s(wl(ELMo, GloVe), tBERT)& 0.743 & 0.667 & 0.618 & 0.703\\
\cline{2-6}
&s(wc(ELMo), wc(GloVe), tBERT)& 0.738 & 0.654 & 0.614 & 0.698\\
\cline{2-6}
&s(wl(ELMo), wl(GloVe), tBERT)& \textbf{0.756} & \textbf{0.684} & \textbf{0.635} & \textbf{0.715}\\
\cline{2-6}
&s(wl(ELMo), wl(GloVe), tBERT, USE)& 0.753 & 0.673 & 0.632 & 0.715\\
\cline{2-6}
&s(wl(ELMo), wl(GloVe), wl(Ling), tBERT)& \textbf{0.753} & \textbf{0.685} & \textbf{0.636} & \textbf{0.718}\\
\cline{2-6}
&s(wc(ELMo), wl(ELMo), wc(GloVe), wl(GloVe), tBERT)& 0.741 & 0.664 & 0.625 & 0.705\\
\hline
\end{tabular}
\end{table}

Table~\ref{results1} shows results produced using traditional ML methods (SVM, RF, and LR) across four different feature sets (word n-grams, character n-grams, averaged ELMo vectors, and composite features). We use Label Powerset for these methods, since the direct (non-transformative) formulation cannot be used with them. Among these combinations, logistic regression with averaged ELMo embeddings as features performs the best. 

\begin{table*}[!t]
\centering
\scriptsize
\caption{Attention analysis for some accounts of sexism}\label{att_ex}
\vspace{-10pt}
\begin{tabular}{|p{3.25in}|p{1.5in}|p{1.25in}|} 
\hline
Account of sexism & Two most-attended-to words / sentence & Labels\\
\hline
\hline
am in social services. my male coworker and I have the same job title. Everyone assumes I am his assistant or an intern. It's not just clients who assume he's my boss, it's other agencies. The same ppl who think nothing calling me honey at work. & (services, am), (coworker, title), (intern, assistant), (boss, agencies), (honey, me) & Role stereotyping, Hostile work environment, Sexual harassment (excluding assault)\\
\hline
I didn't appreciate it when my own father walked into the house one day while I was doing laundry and told me that ``it's nice to see you finally doing women's work.''& (womens, work) & Role stereotyping, Moral policing and victim blaming\\
\hline
being told I should take cat calls as compliments by my father& (compliments, cat) & Denial or trivialisation of sexist misconduct, Sexual harassment (excluding assault)\\
\hline
Referred to as `not a girl' because I have short hair and don't wear noticeable makeup.& (makeup, hair) & Attribute stereotyping, Body shaming\\
\hline
At our school girls are forbidden to wear tight trowsers or remove their blazers- in fear of distracting the boys.& (tight, trowsers) & Moral policing and victim blaming, Hyper-sexualization (excluding body shaming)\\
\hline
\end{tabular}
\end{table*}
Table~\ref{results2} contains results for the random and deep learning baselines and different variants of the proposed framework. For each method, the average over three runs is reported for each metric. We find ELMo to be better than GloVe and fastText for word embeddings across multiple baselines and hence show only ELMO-based results for the baselines. We report all results with the EBCE loss; the NCE loss produced inferior results across multiple methods. For our framework, s() denotes sentence-level concatenation; wl() denotes word level concatenation and LSTM-based processing; wc() denotes word level concatenation and CNN-based processing. We note that the results are reported for only some of the many instances that can arise from our configurable architecture. Our framework provides the ability to explore different configurations such as those with multiple s() operations, depending on the problem at hand. 

We observe the following: (1) The random baseline performs poorly, confirming the complexity of the problem. (2) biLSTM-Attention and Hierarchical-biLSTM-Attention are the two best baselines. (3) Several variants of the proposed framework outperform all baselines. Based on $F_I$ and $F_{macro}$, our best method is s(wl(ELMo), wl(GloVe), tBERT), though adding linguistic features (Ling) to it slightly improves some metrics. (4) BERT tuned on unlabeled instances of sexism (tBERT) works better than the vanilla BERT counterpart and other sentence encoders. (5) Combining tBERT sentence representations with those generated from ELMo word vectors using biLSTM with attention works better than using either individually. (6) Along with tBERT, concatenating ELMo and GloVe at the sentence level (s(wl(ELMo), wl(GloVe), tBERT)) is better than concatenating them at the word level (s(wl(ELMo, GloVe), tBERT)) while processing word vectors using biLSTM with attention. (7) The LSTM based processing of word embeddings produces better results than the CNN based counterpart.

The pre-processing steps that we perform for all deep learning methods include removing certain non-alpha-numeric characters and extra spaces, lower-casing, and zero-padding input tensors as appropriate. While breaking a post into sentences, each sentence containing more than $35$ words is split into multiple sentences, ensuring the maximum sentence length of $35$ words. 

Using experiments on a validation set, which was merged into the training set during the test runs, for each method, we choose the values of three hyper-parameters: the LSTM dimension, the attention dimension, and the number CNN filters for kernel sizes 2, 3, and 4 each. The values used for instances of our framework and the deep learning baselines are provided in Table~\ref{hyper-deep}.

We employ $0.25$ dropouts after each input and before the final, fully connected layer. The learning rate was set to $0.001$ and the number of epochs to $10$. We use a batch size of $64$. These fixed parameters were kept unchanged across methods. 

The hyper-parameter values for the traditional ML methods are as follows. For SVM, soft margin (C) is set to 1.0. For RF (Random Forest), the number of estimators is 100. For extracting character and word n-grams, the maximum number of features used, word n-gram range, and character n-gram range are 10000, (1,2), and (1,5) respectively. For SVM and LR (Logistic Regression), we apply class imbalance correction.

For tapping unlabeled data, we pre-train the `BERT-Base, Uncased' model\footnote{\url{https://github.com/google-research/bert}} for $100000$ steps with a batch size of $25$. For vannila BERT, we use the bigger `BERT-Large, Uncased' model, which we could not use for pre-training because of computational constraints. For generating GloVe word embeddings, we use the 840B-token, cased model.

\begin{table*}[!htb]
\centering
\scriptsize
\caption{Tuned hyper-parameter values for the deep learning baselines (ELMo embeddings) and proposed methods with the EBCE loss}\label{hyper-deep}
\vspace{-10pt}
\begin{tabular}{|p{2.5in}|c|c|c|} 
\hline
Approach & LSTM dimension & Attention dimension & CNN Filters of each kernel size\\
\hline
\hline
Hierarchical-biLSTM-Attention & 300 & 400 & N.A.\\
\hline
CNN-biLSTM-Attention & 300 & 400 & 100\\
\hline
BERT-biLSTM-Attention& 200 & 500 & N.A.\\
\hline
USE-biLSTM-Attention & 300 & 600 & N.A.\\
\hline
InferSent-biLSTM-Attention & 100 & 200 & N.A.\\
\hline
C-biLSTM & 300	& N.A. & 150\\
\hline
biLSTM-Attention & 200 & 300 & N.A.\\
\hline
biLSTM & 300 & N.A. & N.A.\\
\hline
CNN-Kim & N.A. & N.A. & 150\\
\hline
tBERT-biLSTM-Attention & 300 & 600 & N.A.\\
\hline
s(wl(ELMo), tBERT) & 300 & 600 & N.A.\\
\hline
s(wl(ELMo, GloVe), tBERT) & 100 & 100 & N.A.\\
\hline
s(wl(ELMo), wl(GloVe), tBERT) & 100 & 200 & N.A.\\
\hline
s(wl(ELMo), wl(GloVe), tBERT, USE) & 300 & 600 & N.A.\\
\hline
s(wl(ELMo), wl(GloVe), wl(Ling), tBERT) & 300 & 600 & N.A.\\
\hline
s(wc(ELMo), wc(GloVe), tBERT) & 300 & 600 & 100\\
\hline
s(wc(ELMo), wl(ELMo), wc(GloVe), wl(GloVe), tBERT) & 300 & 500 & 100\\
\hline
\end{tabular}
\end{table*}

\begin{table}[!bht]
\centering
\scriptsize
\caption{Performance variation across \#labels per post}\label{num_labs}
\vspace{-10pt}
\begin{tabular}{|l|l|l|l|l|} 
\hline
\#Labels per post & $F_I$ & $F_{macro}$ & $Acc_I$ & $F_{micro}$\\
\hline
\hline
1&0.729&0.527&0.632&0.637\\
\hline
2&0.754&0.675&0.631&0.722\\
\hline
3&0.781&0.721&0.652&0.764\\
\hline
4&0.743&0.722&0.604&0.735\\
\hline
5&0.739&0.592&0.592&0.735\\
\hline
\end{tabular}
\end{table}

Table~\ref{att_ex} lists accounts of sexism from the test set for which our best method made the right predictions but the best baseline did not, along with the labels. It also highlights the top two words per sentence based on the word-level attention weights for wl(ELMo) and wl(GloVe) combined through element-wise max operations. For the first account of sexism, our model produces words like ``intern", ``assistant" and ``boss", associated with role stereotyping, among the top two words across sentences. Likewise, ``honey" related to sexual harassment and ``boss", ``coworker", and ``services" related to hostile work environments also surface. Moreover, the top two sentences based on the sentence-level attention weights of our model are the last two, which evidence all category labels. For other posts too, the model produces category-relevant top two words per sentence; ``womens" and ``work" relate to role stereotyping and moral policing; ``tight" and ``trowsers" relate to hyper-sexualization and moral policing; ``makeup" relates to attribute stereotyping; ``hair" from ``short hair" relates to body shaming; ``cat" from ``cat calls" relates to sexual harassment; ``compliments" from ``take cat calls as compliments" relates to denial or trivialization of sexist misconduct.

Table~\ref{num_labs} shows results for one run of our best method across different numbers of labels per post (1 to 5). Entries for values 6 and 7, which have less than 10 associated test samples, are omitted. The best results are observed for values 2 to 4, suggesting that our approach performs better on multi-label samples.

%% file: conclusion.tex
\section{Conclusion}
\label{sec:conclusion}
We explored classifying an account reporting any kind(s) of sexism such that the categories can co-occur. We developed a neural framework that outperforms many deep learning and traditional ML baselines for this multi-label sexism classification. Moreover, we provided the largest dataset for sexism classification, linked with 23 categories. Directions for future work include devising approaches that perform sexism classification more accurately, enhancing the categorization scheme, and developing other ways to help counter sexism. We hope that this paper will give rise to further work aimed at fighting sexism.

\section{Acknowledgments}
\label{sec:ack}
We are really grateful to Everyday Sexism Project. We also thank Bhavapriya Thottakad, Himakshi Baishya, Aanchal Gupta, Subhashini R, Shalini Pathi, Hadia Ahsan, and Reshma Kalpana, without whose efforts in analyzing accounts of sexism this paper would not have materialized. We are grateful to Ponnurangam Kumaraguru, too, for his time and inputs. 